\documentclass[conference]{IEEEtran}

\usepackage{times}
\usepackage{epsfig}
\usepackage{graphicx}
\usepackage{amsmath}
\usepackage{amssymb}
\usepackage{booktabs}
\usepackage{multicol}
\usepackage{tabularx}
\usepackage{acronym}
\acrodef{SMPL}[SMPL]{\emph{Skinned Multi-Person Linear}}
\acrodef{VIBE}[VIBE]{\emph{Video Inference for Human Body Pose and Shape Estimation}}
\acrodef{MLP}[MLP]{\emph{Multilayer Perceptron}}
\acrodef{CNN}[CNN]{\emph{Convolutional Neural Network}}
\acrodef{FRCNN}[Faster R-CNN]{\emph{Faster Region-based CNN}}
\acrodef{GRU}[GRU]{\emph{Gated Recurrent Unit}}
\acrodef{LDS}[LDS]{\emph{Linear Dynamical Systems}}
\acrodef{mocap}[mocap]{\emph{motion capture}}
\acrodef{HumanID}[USF HumanID]{\emph{USF HumanID Gait Dataset}}
\acrodef{CASIA-B}[CASIA-B]{\emph{CASIA Gait Dataset B}}
\newcommand{\etal}{\textit{et al.}}
\begin{document}

%%%%%%%%% TITLE
\title{Reducing Training Demands for 3D Gait Recognition with\\Deep Koopman Operator Constraints}

\author{Cole Hill, Mauricio Pamplona Segundo, Sudeep Sarkar\\
Computer Science and Engineering, University of South Florida\\
4202 E Fowler Avenue, ENB 118, Tampa, FL 33620\\
{\tt\small \{coleh,mauriciop,sarkar\}@usf.edu}
%\and
}

\maketitle
\thispagestyle{empty}

%%%%%%%%% ABSTRACT
\begin{abstract}
   Deep learning research has made many biometric recognition solution viable, but it requires vast training data to achieve real-world generalization. Unlike other biometric traits, such as face and ear, gait samples cannot be easily crawled from the web to form massive unconstrained datasets. As the human body has been extensively studied for different digital applications, one can rely on prior shape knowledge to overcome data scarcity. This work follows the recent trend of fitting a 3D deformable body model into gait videos using deep neural networks to obtain disentangled shape and pose representations for each frame. To enforce temporal consistency in the network, we introduce a new Linear Dynamical Systems (LDS) module and loss based on Koopman operator theory, which provides an unsupervised motion regularization for the periodic nature of gait, as well as a predictive capacity for extending gait sequences. We compare LDS to the traditional adversarial training approach and use the USF HumanID and CASIA-B datasets to show that LDS can obtain better accuracy with less training data. Finally, we also show that our 3D modeling approach is much better than other 3D gait approaches in overcoming viewpoint variation under normal, bag-carrying and clothing change conditions.
\end{abstract}

\begin{figure*}[h!]
    \centering
    \includegraphics[width=.95\textwidth]{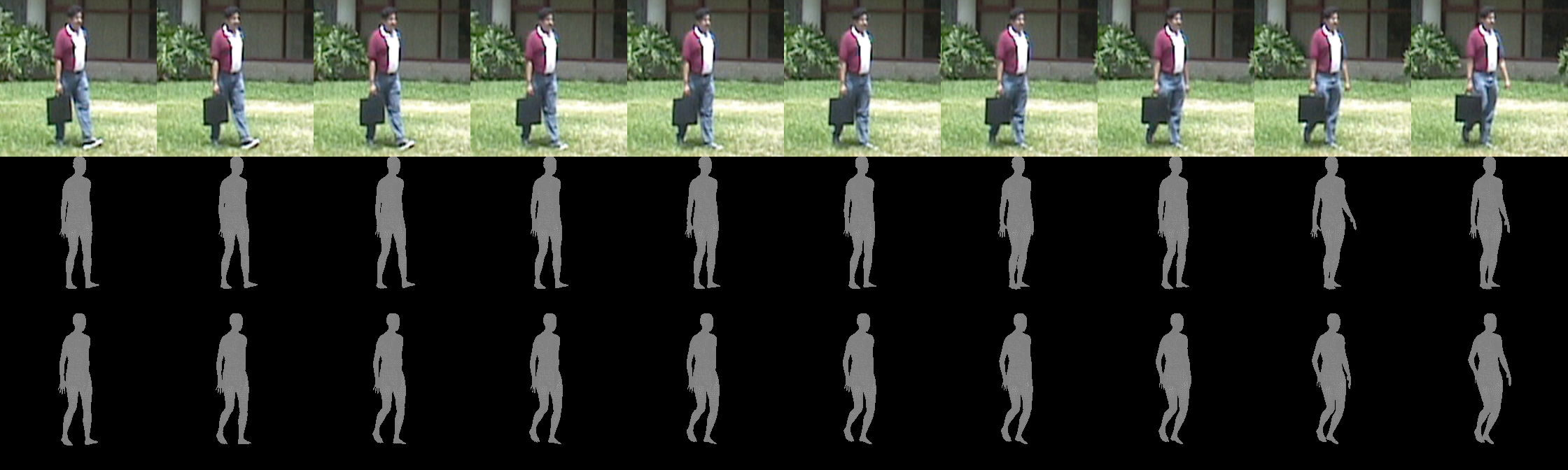}
    \caption{Input sequence from the \acs{HumanID} dataset (top). Predicted SMPL model from VIBE (middle). Predicted SMPL model from LDS embedding (bottom). The LDS predictions are all based on the 10 frames prior to the ones shown, whereas the VIBE model is only using the current frame. Notice that the LDS model captures leg movement when the briefcase occludes the legs were VIBE fails.}
    \label{fig:smpl_predictions}
\end{figure*}

%%%%%%%%% BODY TEXT
\section{Introduction}

Gait recognition is a biometric trait that identifies individuals based on their walking style. Unlike most biometrics, such as fingerprint and iris, gait can be recognized at a distance and without the subject's cooperation, making it valuable for many applications such as security and surveillance. The development of deep learning has spurred recent advances in gait recognition. We can divide the existing works into appearance-based and model-based. Appearance-based approaches~\cite{wu2015learning,shiraga2016geinet,wolf2016multi,lin2021gait,song2019gaitnet,gaitpart,gaitset,lin2021gait} extract a discriminative representation directly from sequences of body silhouette images. Alternatively, model-based approaches~\cite{gaitgraph,liao2017pose,li2020end,zheng2022gait,bansal2022cloth} first fit a human body model (e.g., skeleton, mesh) into RGB images and later derive a discriminative representation from the fitted model parameters.

Unfortunately, the training data requirements for deep learning approaches are immense. While other biometrics have access to massive datasets of annotated images crawled from the internet, like WebFace~\cite{yi2014learning} for face recognition, gait datasets are mostly local data collection research initiatives and thus limited in size. We seek to overcome data scarcity by embedding prior knowledge into the gait recognition model. To do so, we follow recent works~\cite{li2020end,bansal2022cloth} that fit a deformable body model - the \ac{SMPL} model~\cite{loper2015smpl} - to a video sequence and use the model parameters for gait recognition.

Although the \ac{SMPL} model delivers body shape and pose priors, motion coherence over a sequence of frames is not guaranteed. Therefore, we introduce a new \ac{LDS} module to impose temporal consistency and obtain realistic gait sequences. The \ac{LDS} module learns a representation of body pose that follows a linear dynamical system's properties, enforcing periodicity and enabling the prediction of the pose in future frames from previous frames. More importantly, our \ac{LDS} module does not require additional training data or annotations to achieve these outcomes. We contrast \ac{LDS} with the traditional adversarial learning approach in which a discriminator network is trained in parallel to distinguish between real and predicted motion sequences~\cite{kocabas2020vibe}. Unlike \ac{LDS}, the adversarial approach requires real instances of gait motion to perform the same task.

This work contributes to the literature with:
\begin{enumerate}
    \item a novel model-based approach for 3D gait recognition that relies on a disentangled representation of body shape and pose to employ LDS constraints over motion during training;
    \item a new LDS loss that enforces the periodic nature of gait when learning discriminative motion embeddings without requiring additional training data or annotations;
    \item an evaluation of the proposed approach on the \ac{HumanID} and \ac{CASIA-B} highlighting its advantages and showing that it outperforms previous literature works.
\end{enumerate}

\section{Related Works}

\subsection{Gait Recognition}

Recent research in gait recognition has been dominated by approaches utilizing networks trained with deep learning techniques. These works fall primarily in two categories: appearance based and model based. Appearance based approaches derive their gait representations directly from visual representations of gait. Model based approaches first fit a human body model (skeleton or mesh) to the input image frames and then derive features from recognition from the fit model parameters.

Hossaine and Chetty \cite{hossain2013multimodal} showed one of the first appearance based works with deep learning using a Restricted Boltzmann machine (RBF) however results were limited due to the lack of spatial bias in RBF networks. Wu \etal \cite{wu2015learning} and Shiraga \etal \cite{shiraga2016geinet} proposed the first works in gait recognition using the Convolutional Neural Network, whose kernel based structure inherently preserves local spatial relationships. This method used the Gait Energy Image representation first presented in Han and Bhanu \cite{han2005individual}, which compresses the entire gait sequence into a single frame and looses all temporal information. Wolf \etal \cite{wolf2016multi} presented a 3D convolutional approach shortly thereafter which took the silhouettes directly as input and is built from a CNN with spatio-temporal kernels which extract can also extract temporal information. Liao \etal \cite{liao2017pose}. Song \etal \cite{song2019gaitnet} showed the first end-to-end CNN based gait recognition pipeline with GaitNet which extracts silhouettes and discriminative features from RGB imagery with a single network. Chao \etal \cite{gaitset} introduced the set based approach, performing recognition on sets of frames rather than a sequence. Fan \etal \cite{gaitpart} showed performance could be improved by a part based representation which extracted features from horizontal segments of the silhouette. 

Due to the difficulties in extracting accurate 3D information from imagery, model based approaches have been limited to the current state of the art in human body pose and shape models and networks which estimate their parameters from images. Accurate pose estimation was achieved first, Cao \etal \cite{cao2017realtime} enabled Liao \etal \cite{liao2017pose} to create the first model based network which used both a CNN and LSTM to create features for gait recognition. Teepe \etal \cite{gaitgraph} further improved skeleton pose based gait recognition through the use of the graph convolutional network and improved pose estimation with HRNet \cite{sun2019deep}. The release of the Skinned Multi-Person Linear Model (SMPL) by Looper \etal \cite{loper2015smpl} provided a realistic model for both human shape and pose. Li \etal \cite{li2020end} published a end-to-end gait recognition system using SMPL features as input. Zheng \etal \cite{zheng2022gait} introduced a fusion of appearance and model based approaches, extracting appearance based features from silhouettes and normalizing them for pose and view utilizing SMPL parameters. Bansal \etal \cite{bansal2022cloth} integrate SMPL features with the Vision Transformer (ViT) \cite{dosovitskiy2020image} to fuse appearance and model information for recognition. Our proposed is an end-to-end model as in Li \etal however we utilize the newer VIBE model \cite{kocabas2020vibe} for estimating SMPL parameters, as well as the addition of a transformer network for fusion of the temporal pose information and shape parameters.
\begin{figure*}[!h]
  \includegraphics[width=\textwidth]{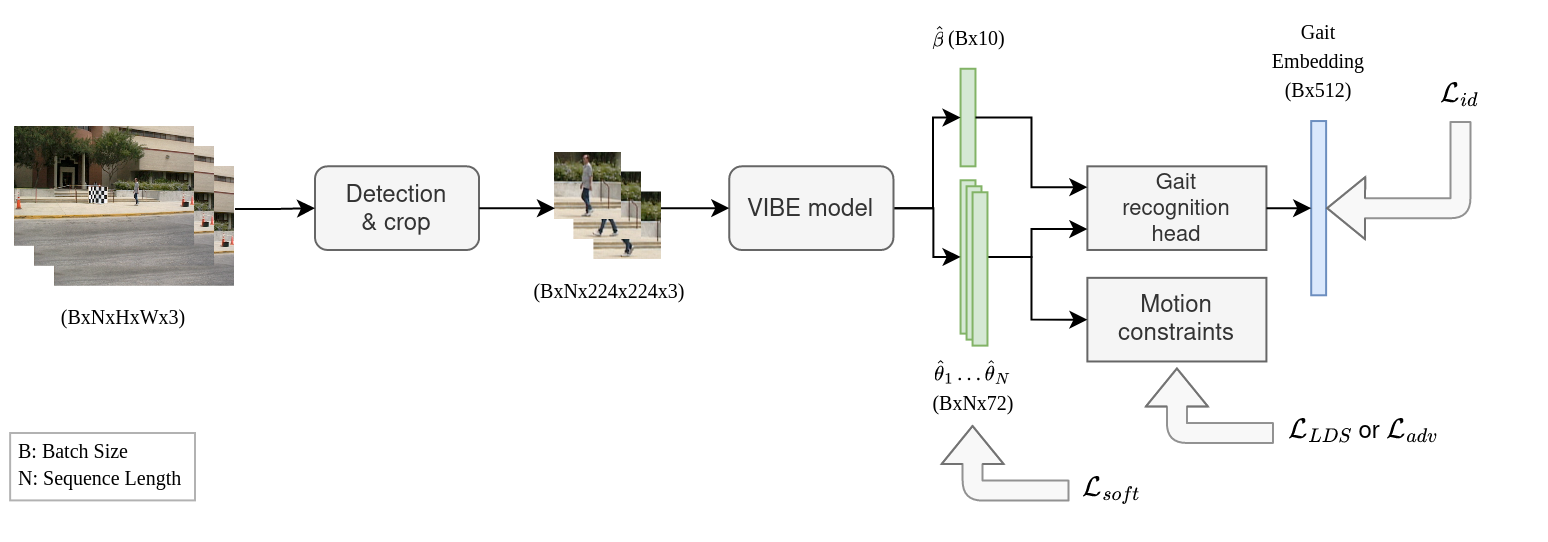}
  \caption{Overview of GaitVIBE and training losses.}
  \label{fig:overview}
\end{figure*}

\subsection{Linear Dynamical Systems in Deep Learning}
Many systems in the real world are nonlinear dynamical systems. There exists methods for approximating these systems as linear such as Dynamic Mode Decomposition \cite{schmid2010dynamic} developed for fluid dynamics. These methods temporally discretize a nonlinear signal in a higher dimensional space called the Koopman Space and estimate an operator called the Koopman Operator which acts as a state transition matrix from one time step to the next in this space. Lusch \etal \cite{lusch2018deep} showed that auto-encoder neural networks can be trained to approximate this behavior if the appropriate constraints and losses are applied. Zhang \etal \cite{zhang2021cross} created a two-part model for silhouette based gait recognition using the principles of the Koopman operator. This work showed promising results, however their model could not be simultaneously optimized for gait recognition and the linear properties required by the Koopman Operator. Additionally their state transition matrix (called a \textit{Koopman} matrix in their paper) was not constrained, making it harder for the model to learn a good representation and making it liable to have numerical stability issues if multiplied many times to forecast far into the future. Our model, inspired by their work, embeds SMPL parameters in a higher dimensional space and enforces they behave as a linear dynamical system. It is a major improvement over prior the prior method as it is a unified model with a single training stage allowing for optimization on multiple tasks simultaneously. Additionally the constrained format of our state transition matrix (a complex diagonal matrix with each entry with unit magnitude) reduces the number of parameters the model needs to learn and guarantees periodic behavior in the latent space when multiplying by the state transition matrix.

\section{Methodology}

\subsection{Architecture}
Our gait recognition architecture is depicted in Figure~\ref{fig:overview}. The recognition process can be divided into three stages: (1) person tracking, (2) 3D body reconstruction, and (3) gait description for matching. We provide details about each of these stages in the following sections.

\subsubsection{Person tracking}
We need to locate the person in each frame to use a video for gait recognition. We start by detecting people in each video frame using the \ac{FRCNN} detector~\cite{methods_fasterrcnn} available in the Detectron2 library~\cite{tools_detectron2}. The outcome of this method is a list of bounding boxes, each enclosing one person in the input frame. Given the controlled nature of the video datasets used in this work, we keep only the largest detection per frame. Although we chose a state-of-the-art detection approach, different factors impact its results for this task: noisy bounding boxes, false detections, missed detections, and multiple people in some frames of some videos. To address those issues, we create three different time series of x, y (center coordinates), and size values from all selected detections of a video. Next, we fit third-degree polynomials to a 150-frame sliding window over these time series with a step size of 50 frames to smooth out the listed problems. Moreover, we take the average of the smoothed values for frames that appear in multiple windows. Finally, we crop the minimum enclosing square around the person's bounding box for the entire track and resize the resulting images to $224\times224$ pixels to form the normalized video that will serve as input to the next stage.

\subsubsection{3D body reconstruction}
We use the \ac{VIBE} architecture~\cite{kocabas2020vibe} unchanged to regress parameters from the \ac{SMPL} model~\cite{loper2015smpl} and thus obtain a 3D body representation for each frame of a normalized video.

The \ac{SMPL} model is a 3D body prior that encodes shape and pose in a disentangled manner. The shape is represented by a 10-dimensional array $\beta$ containing the weights of the ten most significant principal components obtained from thousands of full-body 3D scans. These weights are used to deform a canonical body mesh into any person's physique. The pose is represented by a 72-dimensional array $\theta$, which contains three rotation angles for 23 body joints plus a global orientation. The \ac{SMPL} model receives a pair $(\beta,\theta)$ and produces a mesh with 6890 vertices and 13776 faces. It is worth saying that shape and pose are fully disentangled, so one does not affect the other. 

\ac{VIBE} is a general 3D body fitting approach that retrieves a pair $(\hat\beta_i,\hat\theta_i)$ for the $i$-th video frame. It is composed of a ResNet-50 backbone~\cite{methods_resnet} that extracts relevant visual features from each frame, a 2-layer \ac{GRU}~\cite{methods_gru} that takes care of temporal consistency, and an iterative \ac{MLP}~\cite{methods_mlp} that regresses $\hat\beta_i$ and $\hat\theta_i$ from the temporally adjusted features. During training, \ac{VIBE} relies on body shape and pose supervision (\emph{e.g.}, body joint re-projection error, \ac{SMPL} regression error) and on adversarial motion learning. The latter confronts real examples from a dataset of \ac{mocap} sequences (AMASS~\cite{mahmood2019amass}) with the pose sequences created by \ac{VIBE} to learn how to reproduce realistic motion.

Since the network itself does not impose this, the estimated shape parameter is averaged over the entire sequence to provide a single $\hat\beta$. Figure \ref{fig:smpl_predictions} visualizes some extracted \ac{SMPL} models from RGB images.

\subsubsection{Gait description for matching}
\begin{figure}[!ht]
  \includegraphics[width=\columnwidth]{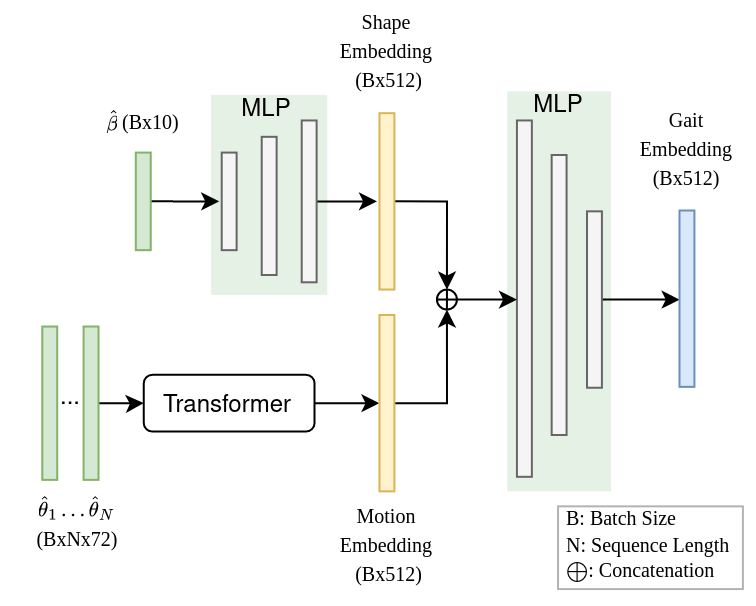}
  \caption{Architecture of the gait recognition head in the GaitVIBE model.}
  \label{fig:gaithead}
\end{figure}

The shape $\hat\beta$ along with the sequence $\hat\theta_1 \dots \hat\theta_N$ of pose parameters for a video of $N$ frames serve as input to the gait recognition head, which maps them into a 512-dimensional embedding that allows us to discriminate between individuals. Its architecture is shown in Figure~\ref{fig:gaithead}.

The $\hat\beta$ parameter is projected from 10 to 512 dimensions using a 2-layer \ac{MLP} with 2048 hidden units. Each layer of the \ac{MLP} consists of a linear layer followed by ReLU activation and batch normalization. We see this network branch as a decoder that extracts the most discerning body measurements from the \ac{SMPL} body representation to form a shape embedding. In parallel, each $\hat\theta_{i}$ parameter is projected into 512 dimensions using a linear layer. We add a special token $[EMB]$ to the beginning of the pose sequence and feed the result to a 6-block Transformer encoder~\cite{methods_transformer}, each block with eight attention heads and 2048 hidden units. We take the output 512-dimensional $[EMB]$ representation as the motion embedding. To produce the final embedding, we concatenate shape and motion embeddings and pass them through another 2-layer \ac{MLP} with 2048 hidden units to produce a single 512-dimensional gait embedding. The output of the last layer goes through an L2 normalization and becomes the final gait embedding. For matching, we compare any two embeddings using the cosine similarity.

\subsection{Loss Functions and Training Details}

\subsubsection{Soft reconstruction loss}
A soft supervision is applied to the \ac{VIBE} backbone by using shape and pose estimates from the original VIBE model as a pseudo ground-truth:
 \begin{equation}
\mathcal{L}_{soft} = ||\beta' - \hat\beta||_2  + \lambda_{pose}\sum_{i=1}^N ||\theta_i' - \hat\theta_i||_2
\end{equation}
with $(\hat\beta,\hat\theta_1\dots\hat\theta_N)$ being the network output, $(\beta',\theta_1'\dots\theta_N')$ the pseudo ground-truth, and $\lambda_{pose}$ a scaling factor to balance the importance of shape and pose ($\lambda_{pose}$ was empirically set to $1000$).  This loss ensures that the model retains valid \ac{SMPL} representations of shape and pose, as the gait datasets used in our experiments (\ac{HumanID} and \ac{CASIA-B}) do not have any kind of annotation other than the person identity.

\subsubsection{Identity loss}
To bias the gait recognition head to use both shape and motion when generating the discriminative embedding, we apply the triplet loss~\cite{methods_tripletloss} to the intermediate shape and motion embeddings ($\mathcal{L}_{triplet\_shape}$ and $\mathcal{L}_{triplet\_motion}$, see Figure~\ref{fig:gaithead}) and the combined gait embedding ($\mathcal{L}_{triplet\_gait}$):
\begin{equation}
\mathcal{L}_{id} = \mathcal{L}_{triplet\_shape} + \mathcal{L}_{triplet\_motion} + \mathcal{L}_{triplet\_gait}
\end{equation}
The margin term is set to $1.0$ in the three cases.

\subsubsection{Motion loss}
Enabled by the disentangled representation of shape and motion provided by the SMPL model, we are able to investigate two alternatives to drive our network to produce realistic gait motion patterns: we follow \ac{VIBE}'s strategy of using a supervised gait-specific adversarial loss and introduce a unsupervised \ac{LDS} loss. %[COMMENT ABOUT DISENTANGLED REPRESENTATION THAT MAKES THIS POSSIBLE???]

\subsubsection*{Gait-specific adversarial learning}
We selected 811 walking instances from the 11000+ \ac{mocap} sequences in the AMASS dataset\cite{mahmood2019amass} to serve as the source of real gait motion examples. With that, we use the following loss term to enforce gait-specific motion patterns in the output of our network:
\begin{equation}
\mathcal{L}_{adv} = \mathbb{E}_{\hat\theta_1\dots\hat\theta_N \sim {p_G}} \left[(D(\hat\theta_1\dots\hat\theta_N)-1)^2\right]
\end{equation}
with $p_G$ being the distribution of pose sequences created by our network and $D$ being a motion discriminator with the same architecture used in \ac{VIBE}~\cite{kocabas2020vibe} (a 2-layer \ac{GRU} followed by a 2-layer \ac{MLP}-based self-attention mechanism and a linear layer for classification). The discriminator itself is trained using the following loss term:
\begin{equation}
\begin{aligned}
\mathcal{L}_{disc} = &~\mathbb{E}_{\theta_1\dots\theta_N \sim {p_R}} \left[(D(\theta_1\dots\theta_N)-1)^2\right] + \\
                     &~\mathbb{E}_{\hat\theta_1\dots\hat\theta_N \sim {p_G}} \left[D(\hat\theta_1\dots\hat\theta_N)^2\right] \\
\end{aligned}
\end{equation}
with $p_R$ being the distribution of real pose sequences. The discriminator is updated at every training iteration of the main network with a learning rate of $10^{-4}$.

\subsubsection*{Linear Dynamical Systems module and loss}
\begin{figure}[!ht]
  \centering
  \includegraphics[width=.9\columnwidth]{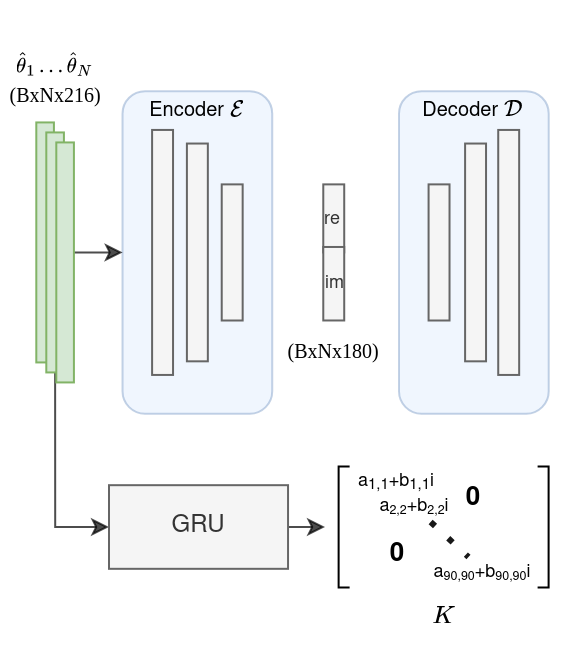}
  \caption{Architecture of the LDS module used for motion constraint.}
  \label{fig:lds}
\end{figure}

The main disadvantage of the adversarial learning strategy for motion constraining is that it requires \ac{mocap} data for training. Knowing that gait data is scarce in the literature, we propose a unsupervised alternative that requires no additional training data but still enforces the periodic nature of gait. Our LDS module is inspired by previous works in learning linear embeddings for nonlinear dynamical systems~\cite{lusch2018deep} and a prior work in gait recognition estimating Koopman matrices and using them as a feature for gait recognition~\cite{zhang2021cross}. The module has two parts, as seen in Figure~\ref{fig:lds}: 1) an autoencoder, and 2) a GRU.

\subsubsection*{Training Details}
We train the 3D reconstruction and gait description modules in an end-to-end fashion. First, we initialize the 3D reconstruction module with the publicly available \ac{VIBE} weights and randomly initialize all other network parameters, except the LDS module, which is bootstrapped by training on the output of the original \ac{VIBE} model. Next, we optimize the entire architecture (except \ac{VIBE}'s ResNet-50 backbone, which is kept frozen) using the Adam algorithm for up to 100 epochs with a learning rate of $5\times10^{-5}$ using a weighted combination of three different losses: soft reconstruction loss $\mathcal{L}_{soft}$, identity loss $\mathcal{L}_{id}$, and motion loss $\mathcal{L}_{motion}$. The model is implemented in Python 3.8 using Pytorch\cite{NEURIPS2019_9015}. 

The autoencoder has an encoder $\mathcal{E}$ and a decoder $\mathcal{D}$, each implemented as a 3-layer \ac{MLP}. Each layer consists of $[216,198,180]$ hidden units respectively and has a ReLU activation function. Before being input to the encoder, the 3 Euler angles representing the orientation of each joint in the SMPL model are converted to $3 \times 3$ rotation matrices. The encoder embeds the sequence of pose parameters into two 90 dimensional vectors representing real and imaginary values in a complex space. The decoder module reconstructs the input sequence from these intermediate vectors. Effectively these create a representation of the gait sequence consisting of 90 sinusoidal signals with variable phases and magnitudes.

The \ac{GRU} takes the first $N/2$ pose parameters in the sequence and produces a single $90\times90$ diagonal matrix $K$ with complex entries. The entries in this matrix all have a unit magnitude. This constraint enforces that when the $K$ matrix is multiplied with one of the latent vectors of the autoencoder, it will only modify the phases of the entries of the vector and not the magnitude. This provides periodicity and numerical stability. The phase of each entry in the $K$ matrix decides the frequency of the 90 latent sinusoids.

The \ac{LDS} loss is the combination of three losses that require no supervision: the autoencoder reconstruction loss $\mathcal{L}_{recons}$, the linearity loss $\mathcal{L}_{linearity}$, and the
recurrent reconstruction $\mathcal{L}_{recons\_rec}$. The reconstruction loss encourages the decoder to reconstruct the pose parameters and is implemented with the standard $SmoothL1Loss$
\begin{equation}
    \mathcal{L}_{recons} = \sum_{i=1}^N ||~\hat\theta_{i} - \mathcal{D}(\mathcal{E}(\hat\theta_{i}))~||_1
\end{equation}
with $||\cdot||_1$ being the smoothed L1 loss implemented in Pytorch \cite{NEURIPS2019_9015}.

The linearity loss optimizes the latent vectors of the autoencoder to be linearly related between time steps. Optimally, $\mathcal{E}(\theta_{i+1})$ should be equal to $K\times \mathcal{E}(\theta_{i})$:
\begin{equation}
    \mathcal{L}_{linearity} = \sum_{i=1}^{N-1} ||~\mathcal{E}(\hat\theta_{i+1}) - K\times\mathcal{E}(\hat\theta_{i})~||_1
\end{equation}

The recurrent reconstruction regularizes the latent vectors and the $K$ matrix such that the entire sequence can be reconstructed by repeatedly multiplying first latent vector by the $K$ matrix:
\begin{equation}
    \mathcal{L}_{recons\_rec} = \sum_{i=1}^{N-1}||~\theta_{i+1},\mathcal{D}(K^{i}\times\mathcal{E}(\theta_{1}))~||_1
\end{equation}

The final \ac{LDS} loss $\mathcal{L}_{LDS}$ is given by: 
\begin{equation}
\mathcal{L}_{LDS} = \mathcal{L}_{recons} + \mathcal{L}_{linearity} + \mathcal{L}_{recons\_rec}
\end{equation}

\subsubsection{Final training loss}

The final training loss is given by:
\begin{equation}
\mathcal{L} = \mathcal{L}_{id} + \lambda_{soft}\mathcal{L}_{soft} + \lambda_{motion}\mathcal{L}_{motion}
\end{equation}
with $\mathcal{L}_{motion}$ being a choice between $\mathcal{L}_{adv}$ or $\mathcal{L}_{LDS}$, and $\lambda_{soft}$ and $\lambda_{motion}$ being scaling factors to balance the three loss terms ($\lambda_{soft}$ was empirically set to $0.06$, and $\lambda_{motion}$ to $0.5$ for $\mathcal{L}_{adv}$ and $1.0$ for $\mathcal{L}_{LDS}$).

\section{Datasets and experiments}

We performed our experiments using the \ac{HumanID}~\cite{sarkar2005humanid} and \ac{CASIA-B}~\cite{yu2006framework}.

The \ac{HumanID} contains gait sequences from 126 subjects outdoors under various covariate conditions, including surface type, time, shoe, briefcase, clothing, and view  (1870 sequences in all). They collected data under the oversight of an IRB, and each captured subject gave informed consent. Each gait sequence is a video taken with one of two outdoor cameras from a person walking on an elliptical path.
The evaluation protocol consists of 12 experiments (Exp. A-L) pairing videos from a fixed gallery set to a probe set with a specific composition of covariates. Only 1080 out of the 1870 videos are used in these experiments. Since this dataset predates the deep learning era, it does not have an official training/test split. As such, we used the gallery set plus all unused videos whose combination of covariates does not appear in any probe set for training (a total of 677 videos).

The \ac{CASIA-B} consists of 124 subjects with ten gait sequences each. These sequences are split into three covariate conditions: 6 normal (NM), 2 carrying a bag (BG), and 2 wearing a coat (CL). They recorded each sequence from 11 view angles. We followed the standard evaluation protocol from previous works, which consists of using the first 74 subjects for training and the rest for testing. At test time, sequences NM \#1-4 compose the gallery, and the remaining ones form three sets of probes: NM \#5-6, BG \#1-2, CL \#1-2.

To further test the robustness of our approach with the inclusion of the \ac{LDS} loss, we also investigate its performance on the \ac{HumanID} after restricting the number of frames in the probe videos.% (details in Section~\ref{sec:restrictedframes}), and after using segments with different trajectories for probe and gallery videos (details in Section~\ref{sec:restrictedsegments}).
In a real-world scenario, a subject may be in view for just a short time before being occluded or walking out of the scene. Therefore a model must be capable of identifying an individual with as few frames as possible. To evaluate the impact of the number of available frames on the \ac{LDS} and adversarial losses, we repeat the \ac{HumanID} experiments after restricting the probe video sizes to a predefined number of frames. Additionally, we exploit the predictive capacity of the \ac{LDS} module to artificially lengthen the input sequences before recognition and investigate when this strategy is beneficial.

%In a real world scenario a subject may be in few for just a short period of time before being occluded or walking out of the scene and therefore it is important that a model is capable of identifying an individual with as little frames as possible. To evaluate how the \ac{LDS} module and adversarial loss affects sensitivity to the number of input frames, the same experiments as before were run, however the number of frames available in the probes was restricted. Both the trained with adversarial learning and \ac{LDS} modules are used. Additionally, the predictive capacity of the \ac{LDS} was used to artificially lengthen the input sequences before they were input to the gait recognition head. 

\begin{table*}
\resizebox{.95\textwidth}{!}{\begin{tabular}{lcccccccccccc}
\textbf{Probe}                                & \multicolumn{1}{c}{A}             & \multicolumn{1}{c}{B}               & \multicolumn{1}{c}{C}               & \multicolumn{1}{c}{D}               & \multicolumn{1}{c}{E}             & F                                   & G                                 & H                                   & I                                   & J                                   & K                                 & L                                   \\ \cline{2-13} 
\multicolumn{1}{l|}{Deng~\emph{et~al.}~\cite{deng2017fusion}} & \multicolumn{1}{c|}{ 97}        & \multicolumn{1}{c|}{\textbf{98}}          & \multicolumn{1}{c|}{96}          & \multicolumn{1}{c|}{92}          & \multicolumn{1}{c|}{88}        & \multicolumn{1}{c|}{81}          & \multicolumn{1}{c|}{82}        & \multicolumn{1}{c|}{95}          & \multicolumn{1}{c|}{92}          & \multicolumn{1}{c|}{82}          & \multicolumn{1}{c|}{76}        & \multicolumn{1}{c|}{73}          \\ \cline{2-13} 
\multicolumn{1}{l|}{Guan~\emph{et~al.}~\cite{guan2015}} & \multicolumn{1}{c|}{\bf 100}        & \multicolumn{1}{c|}{95}          & \multicolumn{1}{c|}{94}          & \multicolumn{1}{c|}{73}          & \multicolumn{1}{c|}{73}        & \multicolumn{1}{c|}{55}          & \multicolumn{1}{c|}{64}        & \multicolumn{1}{c|}{97}          & \multicolumn{1}{c|}{\bf 99}          & \multicolumn{1}{c|}{94}          & \multicolumn{1}{c|}{42}        & \multicolumn{1}{c|}{42}          \\ \cline{2-13} 
\multicolumn{1}{l|}{GaitGraph~\cite{gaitgraph}}        & \multicolumn{1}{c|}{35.20}        & \multicolumn{1}{c|}{74.10}          & \multicolumn{1}{c|}{13.00}          & \multicolumn{1}{c|}{28.80}          & \multicolumn{1}{c|}{19.00}        & \multicolumn{1}{c|}{23.70}          & \multicolumn{1}{c|}{12.10}        & \multicolumn{1}{c|}{63.60}          & \multicolumn{1}{c|}{50.00}          & \multicolumn{1}{c|}{28.00}          & \multicolumn{1}{c|}{63.60}        & \multicolumn{1}{c|}{24.20}          \\ \cline{2-13} 
\multicolumn{1}{l|}{GaitPart~\cite{gaitpart}}          & \multicolumn{1}{c|}{95.90}        & \multicolumn{1}{c|}{81.50}          & \multicolumn{1}{c|}{81.50}          & \multicolumn{1}{c|}{85.60}          & \multicolumn{1}{c|}{77.60}        & \multicolumn{1}{c|}{84.70}          & \multicolumn{1}{c|}{74.10}        & \multicolumn{1}{c|}{89.80}          & \multicolumn{1}{c|}{86.20}          & \multicolumn{1}{c|}{89.00}          & \multicolumn{1}{c|}{84.80}        & \multicolumn{1}{c|}{90.90}          \\ \cline{2-13} 
\multicolumn{1}{l|}{GaitSet~\cite{gaitset}}           & \multicolumn{1}{c|}{\textbf{100}} & \multicolumn{1}{c|}{94.40}          & \multicolumn{1}{c|}{94.40}          & \multicolumn{1}{c|}{93.20}          & \multicolumn{1}{c|}{93.10}        & \multicolumn{1}{c|}{93.20}          & \multicolumn{1}{c|}{89.70}        & \multicolumn{1}{c|}{\textbf{98.30}} & \multicolumn{1}{c|}{93.10}          & \multicolumn{1}{c|}{\textbf{97.50}} & \multicolumn{1}{c|}{81.80}        & \multicolumn{1}{c|}{90.90}          \\ \cline{2-13} 
\multicolumn{1}{l|}{GaitVIBE}                     & \multicolumn{1}{c|}{98.36}        & \multicolumn{1}{c|}{92.59}          & \multicolumn{1}{c|}{\textbf{98.15}} & \multicolumn{1}{c|}{96.61}          & \multicolumn{1}{c|}{96.55}        & \multicolumn{1}{c|}{94.07}          & \multicolumn{1}{c|}{98.28}        & \multicolumn{1}{c|}{97.46}          & \multicolumn{1}{c|}{94.83}          & \multicolumn{1}{c|}{95.76}          & \multicolumn{1}{c|}{66.66}        & \multicolumn{1}{c|}{90.90}          \\ \cline{2-13} 
\multicolumn{1}{l|}{GaitVIBE + Adv}               & \multicolumn{1}{c|}{98.36}        & \multicolumn{1}{c|}{94.44}          & \multicolumn{1}{c|}{\textbf{98.15}} & \multicolumn{1}{c|}{96.61}          & \multicolumn{1}{c|}{98.28}        & \multicolumn{1}{c|}{94.07}          & \multicolumn{1}{c|}{96.55}        & \multicolumn{1}{c|}{94.07}          & \multicolumn{1}{c|}{94.83}          & \multicolumn{1}{c|}{94.92}          & \multicolumn{1}{c|}{75.76}        & \multicolumn{1}{c|}{90.91}          \\ \cline{2-13} 
\multicolumn{1}{l|}{GaitVIBE + LDS}               & \multicolumn{1}{c|}{\textbf{100}} & \multicolumn{1}{c|}{96.30} & \multicolumn{1}{c|}{\textbf{98.15}} & \multicolumn{1}{c|}{\textbf{97.46}} & \multicolumn{1}{c|}{\textbf{100}} & \multicolumn{1}{c|}{\textbf{95.76}} & \multicolumn{1}{c|}{\textbf{100}} & \multicolumn{1}{c|}{95.76}          & \multicolumn{1}{c|}{98.28} & \multicolumn{1}{c|}{96.61}          & \multicolumn{1}{c|}{\textbf{100}} & \multicolumn{1}{c|}{\textbf{96.97}} \\ \cline{2-13} 
\end{tabular}} % Rank-1 Results vs other methods
\newline
\caption{Rank-1 Accuracies of the stated method vs. prior methods on the \ac{HumanID}. The top accuracy for each experiment category is indicated in bold.}
\label{tab:humanid_rank1}
\end{table*}

\begin{table*}
\resizebox{.95\textwidth}{!}{\begin{tabular}{lcccccccccccc}
\textbf{Probe}                                & \multicolumn{1}{c}{A}             & \multicolumn{1}{c}{B}             & \multicolumn{1}{c}{C}             & \multicolumn{1}{c}{D}             & \multicolumn{1}{c}{E}             & F                                 & G                                 & H                                 & I                                 & J                                 & K                                 & L                                 \\ \cline{2-13} 
\multicolumn{1}{l|}{Deng~\emph{et~al.}~\cite{deng2017fusion}} & \multicolumn{1}{c|}{ \textbf{100}}        & \multicolumn{1}{c|}{\textbf{100}}          & \multicolumn{1}{c|}{\textbf{100}}          & \multicolumn{1}{c|}{98}          & \multicolumn{1}{c|}{95}        & \multicolumn{1}{c|}{88}          & \multicolumn{1}{c|}{85}        & \multicolumn{1}{c|}{98}          & \multicolumn{1}{c|}{\textbf{100}}          & \multicolumn{1}{c|}{92}          & \multicolumn{1}{c|}{82}        & \multicolumn{1}{c|}{79}          \\ \cline{2-13}
\multicolumn{1}{l|}{Guan~\emph{et~al.}~\cite{guan2015}} & \multicolumn{1}{c|}{\bf 100}        & \multicolumn{1}{c|}{98}          & \multicolumn{1}{c|}{98}          & \multicolumn{1}{c|}{85}          & \multicolumn{1}{c|}{84}        & \multicolumn{1}{c|}{73}          & \multicolumn{1}{c|}{79}        & \multicolumn{1}{c|}{98}          & \multicolumn{1}{c|}{99}          & \multicolumn{1}{c|}{98}          & \multicolumn{1}{c|}{55}        & \multicolumn{1}{c|}{58}          \\ \cline{2-13} 
\multicolumn{1}{l|}{GaitGraph~\cite{gaitgraph}}        & \multicolumn{1}{c|}{78.70}        & \multicolumn{1}{c|}{83.30}        & \multicolumn{1}{c|}{57.40}        & \multicolumn{1}{c|}{55.90}        & \multicolumn{1}{c|}{41.40}        & \multicolumn{1}{c|}{55.10}        & \multicolumn{1}{c|}{44.80}        & \multicolumn{1}{c|}{83.10}        & \multicolumn{1}{c|}{69.00}        & \multicolumn{1}{c|}{67.80}        & \multicolumn{1}{c|}{72.70}        & \multicolumn{1}{c|}{51.50}        \\ \cline{2-13} 
\multicolumn{1}{l|}{GaitPart~\cite{gaitpart}}          & \multicolumn{1}{c|}{99.20}        & \multicolumn{1}{c|}{88.90}        & \multicolumn{1}{c|}{92.60}        & \multicolumn{1}{c|}{91.50}        & \multicolumn{1}{c|}{94.80}        & \multicolumn{1}{c|}{94.10}        & \multicolumn{1}{c|}{91.40}        & \multicolumn{1}{c|}{95.80}        & \multicolumn{1}{c|}{96.60}        & \multicolumn{1}{c|}{96.60}        & \multicolumn{1}{c|}{93.90}        & \multicolumn{1}{c|}{97.00}        \\ \cline{2-13} 
\multicolumn{1}{l|}{GaitSet~\cite{gaitset}}           & \multicolumn{1}{c|}{\textbf{100}} & \multicolumn{1}{c|}{96.30}        & \multicolumn{1}{c|}{\textbf{100}} & \multicolumn{1}{c|}{98.30}        & \multicolumn{1}{c|}{\textbf{100}} & \multicolumn{1}{c|}{99.20}        & \multicolumn{1}{c|}{94.80}        & \multicolumn{1}{c|}{\textbf{100}} & \multicolumn{1}{c|}{\textbf{100}} & \multicolumn{1}{c|}{\textbf{100}} & \multicolumn{1}{c|}{97.00}        & \multicolumn{1}{c|}{93.90}        \\ \cline{2-13} 
\multicolumn{1}{l|}{GaitVIBE}                     & \multicolumn{1}{c|}{\textbf{100}} & \multicolumn{1}{c|}{98.15}        & \multicolumn{1}{c|}{\textbf{100}} & \multicolumn{1}{c|}{\textbf{100}} & \multicolumn{1}{c|}{\textbf{100}} & \multicolumn{1}{c|}{\textbf{100}} & \multicolumn{1}{c|}{\textbf{100}} & \multicolumn{1}{c|}{\textbf{100}} & \multicolumn{1}{c|}{98.28}        & \multicolumn{1}{c|}{99.15}        & \multicolumn{1}{c|}{87.88}        & \multicolumn{1}{c|}{96.97}        \\ \cline{2-13} 
\multicolumn{1}{l|}{GaitVIBE + Adv}               & \multicolumn{1}{c|}{\textbf{100}} & \multicolumn{1}{c|}{\textbf{100}} & \multicolumn{1}{c|}{\textbf{100}} & \multicolumn{1}{c|}{\textbf{100}} & \multicolumn{1}{c|}{\textbf{100}} & \multicolumn{1}{c|}{\textbf{100}} & \multicolumn{1}{c|}{\textbf{100}} & \multicolumn{1}{c|}{\textbf{100}} & \multicolumn{1}{c|}{98.28}        & \multicolumn{1}{c|}{99.15}        & \multicolumn{1}{c|}{93.94}        & \multicolumn{1}{c|}{\textbf{100}} \\ \cline{2-13} 
\multicolumn{1}{l|}{GaitVIBE + LDS}               & \multicolumn{1}{c|}{\textbf{100}} & \multicolumn{1}{c|}{\textbf{100}} & \multicolumn{1}{c|}{\textbf{100}} & \multicolumn{1}{c|}{\textbf{100}} & \multicolumn{1}{c|}{\textbf{100}} & \multicolumn{1}{c|}{99.15}        & \multicolumn{1}{c|}{\textbf{100}} & \multicolumn{1}{c|}{99.15}        & \multicolumn{1}{c|}{\textbf{100}} & \multicolumn{1}{c|}{99.15}        & \multicolumn{1}{c|}{\textbf{100}} & \multicolumn{1}{c|}{\textbf{100}} \\ \cline{2-13} 
\end{tabular}} % Rank-5 Results vs other methods
\newline
\caption{Rank-5 Accuracies of the stated method vs. prior methods on the \ac{HumanID}. The top accuracy for each experiment category is indicated in bold.}
\label{tab:humanid_rank5}
\end{table*}

%\subsection{Restricting part of sequence in Probe/Gallery}
%\label{sec:restrictedsegments}
%
%To further test the model under viewpoint conditions beyond the two cameras available in the USF HumanID dataset, additional experiments where run by changing the part of the sequence available in the Gallery and Probes at test time. The elliptical path walked by subjects in the USF HumanID dataset can be split into 3 conditions (near, far, and turning). Each of the sequences were split and classified into these three classes. Each of these classes, with the addition of all frames were tested as gallery and probe using the previously described experimental regime. For example using the entire sequence in the gallery but only the turning part of the sequence in the probe.

%\section{Results}
\subsection{\ac{HumanID} results}

Tables~\ref{tab:humanid_rank1}~and~\ref{tab:humanid_rank5} show the Rank-1 and Rank-5 accuracy of different approaches in the \ac{HumanID}. \emph{GaitVIBE} stands for the results of the proposed model without any motion constraints, while \emph{GaitVIBE + Adv} includes the adversarial learning strategy and \emph{GaitVIBE + LDS} the proposed \ac{LDS} module. In addition, to provide a point of reference, we include the best \ac{HumanID} performance reported in the literature and the results of three state-of-the-art approaches, GaitGraph~\cite{gaitgraph}, GaitPart~\cite{gaitpart}, and GaitSet~\cite{gaitset}. We used their implementations from the OpenGait repository to train these approaches using our \ac{HumanID} setup. GaitPart and GaitSet require body silhouette images, so we use a background matting method~\cite{method_bgmatting} that segments foreground objects to extract those.

Table~\ref{tab:humanid_rank1} shows that the proposed model matches or outperforms prior works in almost all probe sets regarding Rank-1 performance, except for Experiments H-J. Notably, all of these experiments are the briefcase-carrying condition indicating that the VIBE backbone may have issues with leg occlusion. Figure~\ref{fig:smpl_predictions} shows this effect in the fifth frame where the leg is bent, but the estimated \ac{SMPL} pose has both feet on the ground. Additionally, we see that even though \emph{GaitVIBE + LDS} requires less training data than \emph{GaitVIBE + Adv}, it was the top performer. Compared to \emph{GaitVIBE}, which uses the same amount of training data and has no motion constraints, the boost in performance given by \ac{LDS} is even more evident. Further, both \emph{GaitSet}\cite{gaitset} and \emph{GaitPart}\cite{gaitpart} show drops in performance for Probes B-L where the ground surface or shoe was changed, especially in probes E and G where both are changed. This indicates silhouette based models may lend too much weight to small irrelevant details in the silhouette such as shoe shape or a shadow artefact from the silhouette extraction. Our whole body representation and disentagled pose force the model to be more robust to these changes. Similar observations can be made in Table~\ref{tab:humanid_rank5} for the Rank-5 results, showing the benefit of the LDS loss for gait recognition.
%[COMMENT ON THE ADVANTAGE OF THIS APPROACH OVER SILHOUETTE-BASED METHODS FOR THIS SPECIFIC DATASET???]

\begin{figure}[!ht]
    \centering
    \includegraphics[width=\columnwidth]{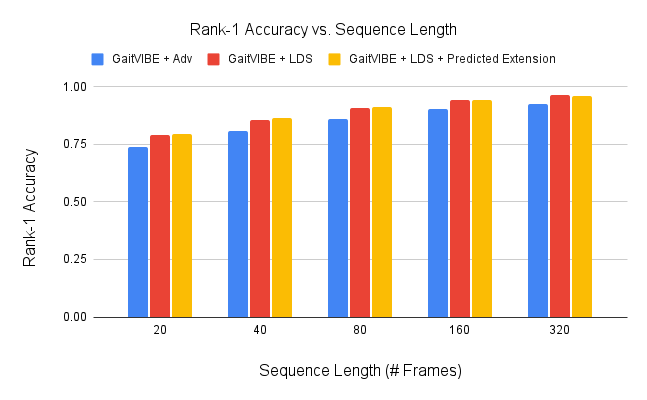}
    \caption{Rank-1 Accuracy vs. The number of input frames for the proposed gait recognition model. The blue bars represent the model using an adversarial loss discriminating between synthetic sequences and sequences in the AMASS dataset. The red bars represent the model without adversarial loss, but instead the LDS loss and additional triplet losses. Finally the yellow bars show the LDS model with the sequences extended by 40 frames using the LDS embeddings.}
    \label{fig:rank1vsseq}
\end{figure}

Figure~\ref{fig:rank1vsseq} shows the degradation in Rank-1 performance as we reduce the number of frames in the probe videos. Here again, the benefit of the \ac{LDS} loss is shown as \emph{GaitVIBE + LDS} can outperform \emph{GaitVIBE + Adv} by a growing margin as the number of frames decreases. Additionally, this graph shows that we can get a slight gain in performance by using the predictive capacity of the \ac{LDS} latent space to lengthen the input sequence. Here we estimate the matrix $K$ using the input sequence and then infer 40 additional poses by multiplying this matrix by the final pose embedding and decoding it using the decoder $\mathcal{D}$. The gains are enumerated in Table~\ref{table:ext_vs_noext}.

\begin{table}[!ht]
\resizebox{.46\textwidth}{!}{\begin{tabular}{lrrrrr}
\textbf{Sequence Length}                        & 20                       & 40                       & 80                       & 160                      & 320                      \\ \cline{2-6} 
\multicolumn{1}{l|}{GaitVIBE + LDS}                 & \multicolumn{1}{r|}{6.9} & \multicolumn{1}{r|}{5.9} & \multicolumn{1}{r|}{5.9} & \multicolumn{1}{r|}{4.3} & \multicolumn{1}{r|}{4.2} \\ \cline{2-6} 
\multicolumn{1}{l|}{GaitVIBE + LDS + Predicted Ext} & \multicolumn{1}{r|}{7.9} & \multicolumn{1}{r|}{6.7} & \multicolumn{1}{r|}{6.5} & \multicolumn{1}{r|}{4.6} & \multicolumn{1}{r|}{4.1} \\ \cline{2-6} 
\end{tabular}}
\newline
\caption{Gain in rank-1 accuracy using linear dynamical loss and extension of gait sequence compared with VIBE with adversarial loss.}
\label{table:ext_vs_noext}
\end{table}

%\begin{figure}
%    \centering
%    \includegraphics[width=\columnwidth]{Rank-1 Accuracy gain for VIBE + LDS.png}
%    \caption{Gain in Rank-1 Accuracy when using VIBE model with LDS loss over VIBE model with adversarial loss for viewpoint variation experiments. The labels on the x-axis indicate the split that was used for example \textbf{All vs. Turn} indicates that the entire sequence was used in the gallery, however only the turns were used in the Probe.}
%    \label{fig:rank1vsview}
%\end{figure}

\begin{table*}[!ht]
\resizebox{.95\textwidth}{!}{

\begin{tabular}{llcccccccccccc}
                        & \textbf{View Angle}             & 0                                   & 18                                  & 36                                  & 54                                  & 72                                  & 90                                  & 108                                 & 126                                 & 144                                 & 162                                 & 180                                 &   Avg                 \\ \cline{2-14} 
\multicolumn{1}{l|}{}   & \multicolumn{1}{l|}{GaitVIBE}       & \multicolumn{1}{r|}{60.80}          & \multicolumn{1}{r|}{66.20}          & \multicolumn{1}{r|}{64.70}          & \multicolumn{1}{r|}{59.60}          & \multicolumn{1}{r|}{56.30}          & \multicolumn{1}{r|}{51.90}          & \multicolumn{1}{r|}{60.10}          & \multicolumn{1}{r|}{65.00}          & \multicolumn{1}{r|}{61.90}          & \multicolumn{1}{r|}{64.00}          & \multicolumn{1}{r|}{58.70}          & \multicolumn{1}{r|}{60.84}          \\ \cline{3-14} 
\multicolumn{1}{l|}{NM} & \multicolumn{1}{l|}{GaitVIBE + Adv} & \multicolumn{1}{r|}{\textbf{64.30}} & \multicolumn{1}{r|}{\textbf{69.90}} & \multicolumn{1}{r|}{\textbf{70.80}} & \multicolumn{1}{r|}{\textbf{69.40}} & \multicolumn{1}{r|}{\textbf{65.10}} & \multicolumn{1}{r|}{\textbf{66.10}} & \multicolumn{1}{r|}{\textbf{70.40}} & \multicolumn{1}{r|}{\textbf{67.20}} & \multicolumn{1}{r|}{\textbf{67.20}} & \multicolumn{1}{r|}{\textbf{65.20}} & \multicolumn{1}{r|}{59.30}          & \multicolumn{1}{r|}{\textbf{66.81}} \\ \cline{3-14} 
\multicolumn{1}{l|}{}   & \multicolumn{1}{l|}{GaitVIBE + LDS} & \multicolumn{1}{r|}{63.40}          & \multicolumn{1}{r|}{68.20}          & \multicolumn{1}{r|}{70.60}          & \multicolumn{1}{r|}{66.20}          & \multicolumn{1}{r|}{60.10}          & \multicolumn{1}{r|}{56.00}          & \multicolumn{1}{r|}{65.20}          & \multicolumn{1}{r|}{66.20}          & \multicolumn{1}{r|}{62.50}          & \multicolumn{1}{r|}{61.10}          & \multicolumn{1}{r|}{\textbf{62.10}} & \multicolumn{1}{r|}{63.78}          \\ \cline{2-14} 
\multicolumn{1}{l|}{}   & \multicolumn{1}{l|}{GaitVIBE}       & \multicolumn{1}{r|}{45.20}          & \multicolumn{1}{r|}{48.30}          & \multicolumn{1}{r|}{46.60}          & \multicolumn{1}{r|}{40.00}          & \multicolumn{1}{r|}{38.10}          & \multicolumn{1}{r|}{40.40}          & \multicolumn{1}{r|}{43.20}          & \multicolumn{1}{r|}{45.20}          & \multicolumn{1}{r|}{\textbf{49.30}} & \multicolumn{1}{r|}{45.50}          & \multicolumn{1}{r|}{42.40}          & \multicolumn{1}{r|}{44.02}          \\ \cline{3-14} 
\multicolumn{1}{l|}{BG} & \multicolumn{1}{l|}{GaitVIBE + Adv} & \multicolumn{1}{r|}{\textbf{55.10}} & \multicolumn{1}{r|}{\textbf{57.50}} & \multicolumn{1}{r|}{\textbf{57.20}} & \multicolumn{1}{r|}{\textbf{49.10}} & \multicolumn{1}{r|}{\textbf{39.80}} & \multicolumn{1}{r|}{\textbf{46.00}} & \multicolumn{1}{r|}{\textbf{48.20}} & \multicolumn{1}{r|}{\textbf{46.00}} & \multicolumn{1}{r|}{48.90}          & \multicolumn{1}{r|}{\textbf{47.40}} & \multicolumn{1}{r|}{\textbf{46.50}} & \multicolumn{1}{r|}{\textbf{49.25}} \\ \cline{3-14} 
\multicolumn{1}{l|}{}   & \multicolumn{1}{l|}{GaitVIBE + LDS} & \multicolumn{1}{c|}{48.40}          & \multicolumn{1}{c|}{53.10}          & \multicolumn{1}{c|}{49.20}          & \multicolumn{1}{c|}{43.70}          & \multicolumn{1}{c|}{37.50}          & \multicolumn{1}{c|}{36.10}          & \multicolumn{1}{c|}{34.40}          & \multicolumn{1}{c|}{37.90}          & \multicolumn{1}{c|}{43.70}          & \multicolumn{1}{c|}{44.40}          & \multicolumn{1}{c|}{44.20}          & \multicolumn{1}{r|}{42.96}          \\ \cline{2-14} 
\multicolumn{1}{l|}{}   & \multicolumn{1}{l|}{GaitVIBE}       & \multicolumn{1}{r|}{21.20}          & \multicolumn{1}{r|}{\textbf{29.50}} & \multicolumn{1}{r|}{26.20}          & \multicolumn{1}{r|}{23.50}          & \multicolumn{1}{r|}{\textbf{19.50}} & \multicolumn{1}{r|}{\textbf{20.70}} & \multicolumn{1}{r|}{\textbf{18.80}} & \multicolumn{1}{r|}{\textbf{21.70}} & \multicolumn{1}{r|}{\textbf{20.70}} & \multicolumn{1}{r|}{19.00}          & \multicolumn{1}{r|}{19.90}          & \multicolumn{1}{r|}{\textbf{21.88}} \\ \cline{3-14} 
\multicolumn{1}{l|}{CL} & \multicolumn{1}{l|}{GaitVIBE + Adv} & \multicolumn{1}{r|}{\textbf{25.20}} & \multicolumn{1}{r|}{28.40}          & \multicolumn{1}{r|}{24.60}          & \multicolumn{1}{r|}{20.20}          & \multicolumn{1}{r|}{17.70}          & \multicolumn{1}{r|}{19.30}          & \multicolumn{1}{r|}{18.40}          & \multicolumn{1}{r|}{20.10}          & \multicolumn{1}{r|}{16.30}          & \multicolumn{1}{r|}{\textbf{19.80}} & \multicolumn{1}{r|}{\textbf{20.10}} & \multicolumn{1}{r|}{20.92}          \\ \cline{3-14} 
\multicolumn{1}{l|}{}   & \multicolumn{1}{l|}{GaitVIBE + LDS} & \multicolumn{1}{c|}{20.60}          & \multicolumn{1}{c|}{28.10}          & \multicolumn{1}{c|}{\textbf{27.60}} & \multicolumn{1}{c|}{\textbf{23.70}} & \multicolumn{1}{c|}{16.40}          & \multicolumn{1}{c|}{17.10}          & \multicolumn{1}{c|}{16.30}          & \multicolumn{1}{c|}{18.80}          & \multicolumn{1}{c|}{18.60}          & \multicolumn{1}{c|}{17.80}          & \multicolumn{1}{c|}{17.30}          & \multicolumn{1}{r|}{20.21}          \\ \cline{2-14} 
\end{tabular}

} % Rank-1 Results vs other methods
\newline
\caption{Rank-1 Accuracies of the stated method on \ac{CASIA-B}. The top accuracy for each experiment category is indicated in bold.}
\label{casia_table}
\end{table*}

\begin{table}[]
    \centering
    \resizebox{\columnwidth}{!}{
\begin{tabular}{llrrr}
 & \textbf{} & \multicolumn{1}{l}{NM} & \multicolumn{1}{l}{BG} & \multicolumn{1}{l}{CL} \\ \cline{1-5}
\multicolumn{1}{|c}{Appearance} &
 \multicolumn{1}{|l|}{GaitNet \cite{song2019gaitnet}} & \multicolumn{1}{r|}{91.6} & \multicolumn{1}{r|}{85.7} & \multicolumn{1}{r|}{58.9} \\ \cline{2-5}
 \multicolumn{1}{|c}{Based}&
 \multicolumn{1}{|l|}{GaitSet \cite{gaitset}} & \multicolumn{1}{r|}{95.0} & \multicolumn{1}{r|}{87.2} & \multicolumn{1}{r|}{70.4} \\ \cline{2-5}
\multicolumn{1}{|c}{Approaches} &
 \multicolumn{1}{|l|}{GaitPart \cite{gaitpart}} & \multicolumn{1}{r|}{96.2} & \multicolumn{1}{r|}{91.5} & \multicolumn{1}{r|}{78.7} \\ \cline{2-5}
 \multicolumn{1}{|c}{} &
 \multicolumn{1}{|l|}{GaitGL \cite{lin2021gait}} & \multicolumn{1}{r|}{\textbf{97.4}} & \multicolumn{1}{r|}{\textbf{94.5}} & \multicolumn{1}{r|}{\textbf{83.6}} \\ \cline{1-5}
 \multicolumn{1}{|c}{3D Model}&
 \multicolumn{1}{|l|}{PoseGait \cite{liao2020model}} & \multicolumn{1}{r|}{63.78} & \multicolumn{1}{r|}{45.52} & \multicolumn{1}{r|}{31.98} \\ \cline{2-5}
\multicolumn{1}{|c}{Based}& 
\multicolumn{1}{|l|}{Li \etal\cite{li2022multi}} & \multicolumn{1}{r|}{76.64} & \multicolumn{1}{r|}{42.01} & \multicolumn{1}{r|}{32.81} \\ \cline{2-5}
\multicolumn{1}{|c}{Approaches}& 
\multicolumn{1}{|l|}{GaitVIBE} & \multicolumn{1}{r|}{88.53} & \multicolumn{1}{r|}{79.72} & \multicolumn{1}{r|}{51.57} \\ \cline{2-5}
\multicolumn{1}{|c}{}& 
\multicolumn{1}{|l|}{GaitVIBE + Adv} & \multicolumn{1}{r|}{\textbf{92.61}} & \multicolumn{1}{r|}{\textbf{87.54}} & \multicolumn{1}{r|}{54.08} \\ \cline{2-5}
\multicolumn{1}{|c}{}& 
\multicolumn{1}{|l|}{GaitVIBE + LDS} & \multicolumn{1}{r|}{90.05} & \multicolumn{1}{r|}{83.81} & \multicolumn{1}{r|}{\textbf{55.50}} \\ \cline{1-5}
\end{tabular}
}
    \caption{Comparison of stated work with prior appearance and 3D gait recognition based approaches on \ac{CASIA-B}. Best ranking results for each category is indicated in bold.}
    \label{tab:casia_comp}
\end{table}

\subsection{\ac{CASIA-B} results}

Table \ref{casia_table} shows the Rank-1 performance of the stated model with the 3 loss conditions on the \ac{CASIA-B} dataset. Here the CNN backbone of the VIBE model is frozen during training. We see that the LDS loss provides less of a benefit than, but close to the adversarial case. This may be due to the different walking conditions of the two datasets. \ac{HumanID} has longer gait sequences in a circular path which may be more periodic than the straight line walked in the \ac{CASIA-B} dataset; the gait changes as the subject approaches the end points. This may diminish the benefit of the periodic prior built into the LDS module. The LDS loss still outperforms the plain model in all cases for the NM set. Results are mixed in the BG and CL sets, with the adversarial loss still showing a benefit in the BG set but no clear benefit from the LDS module in BG. All versions of the model gave similar performance in the CL set. 

Table \ref{tab:casia_comp} compares the results of our model with other 3D gait recognition methods on the \ac{CASIA-B} dataset. To enable a fair comparison with Li \etal~\cite{li2020end} in this experiment we have trained the entire VIBE model including the CNN backbone. In their work they also train the entire network and use the AMASS dataset for adversarial learning. Results indicate that all three of versions of our models outperform other 3D model based approaches reported on the \ac{CASIA-B} dataset. As with our other results on \ac{CASIA-B} our \emph{GaitVIBE + Adv} outperforms \emph{GaitVIBE} and \emph{GaitVIBE + LDS} for the NM and BG sets. However, \emph{GaitVIBE + LDS} showed best performance on the CL set and improves upon \emph{GaitVIBE} on all sets without the additional AMASS dataset required for adversarial learning.

\section{Societal Impact}

Gait recognition as a technology has wide ranging applications in many aspects of human lives, and brings forward ethical considerations about how the technology is used and by who. There are concerns about the invasion of individuals' privacy and the tracking of their location. The growing number of public cameras exacerbates the issue with most public places being recorded. It is important for the research community to continue to insist on IRB oversight and informed consent for individuals who are subjects in this research, especially for "in-the-wild" collections. The USF HumanID dataset used in this paper, was collected in controlled laboratory conditions using an IRB approved informed consent process. 

There are, however, many social goods that could come from the ability to identify individuals at a distance. Potential applications range from remotely identifying poachers of endangered species to the location of lost mentally impaired individuals or children. It could even help in day to day lives by enabling smart stores with no checkout. 

\section{Conclusion}

In this paper we present a gait recognition method utilizing 3D human body models with linear dynamical systems constraints via Koopman operator theory. Results show that the model is superior to prior works on the \ac{HumanID} and 3D gait recognition based approaches on the \ac{CASIA-B} datasets. The LDS constraints allow the model to show improved performance without the extra data required for adversarial training. Results on the \ac{CASIA-B} dataset show that these benefits may be reduced under different conditions. We believe that these LDS losses could be applied in the future to other problems where there is underlying periodic nature beyond gait recognition.

\section*{Acknowledgements}
This research is based upon work supported in part by the Office of the Director of National Intelligence (ODNI), Intelligence Advanced Research Projects Activity (IARPA), via 2022-21102100003. The views and conclusions contained herein are those of the authors and should not be interpreted as necessarily representing the official policies, either expressed or implied, of ODNI, IARPA, or the U.S. Government. The U.S. Government is authorized to reproduce and distribute reprints for governmental purposes notwithstanding any copyright annotation therein.

{\small
\bibliographystyle{ieee_fullname}
\bibliography{ref}
}

\end{document}